%% file: neurips_2020_tda.tex
\documentclass{article}

\usepackage[final,nonatbib]{neurips_2020_tda}

\usepackage[utf8]{inputenc} 
\usepackage[T1]{fontenc}    
\usepackage{amsfonts}       
\usepackage{booktabs}       
\usepackage[hidelinks]{hyperref}       
\usepackage{url}            
\usepackage{microtype}      
\usepackage{nicefrac}       
\usepackage{paralist}       
\usepackage{siunitx}        
\usepackage[font=small]{caption}
\usepackage[labelformat=empty, position=top]{subcaption}
\usepackage[export]{adjustbox}

\usepackage{amsmath,amssymb,amsfonts,amsthm}
\usepackage{microtype}
\usepackage{algorithmic}
\usepackage{graphicx}
\usepackage{textcomp}
\usepackage{xcolor}
\usepackage{tikz}
\usetikzlibrary{calc}
\usepackage{mwe}
\usepackage{blkarray}

\usepackage{dcolumn,booktabs}
\newcolumntype{d}[1]{D{.}{.}{#1}}

\newsavebox{\tempbox}
\newlength{\tempwidth}

\usepackage[natbib,hyperref,sorting=none]{biblatex}
\title{
  Simplicial Neural Networks
}

\author{%
  Stefania Ebli\\
  Laboratory for Topology and Neuroscience\\
  EPFL, Lausanne, Switzerland\\
  \texttt{stefania.ebli@epfl.ch}
  \And
  Michaël Defferrard\\
  Signal Processing Laboratory (LTS2)\\
  EPFL, Lausanne, Switzerland\\
  \texttt{michael.defferrard@epfl.ch}
  \And
  Gard Spreemann\thanks{Work done while at the EPFL. Present affiliation: Telenor Research, Fornebu, Norway.}\\
  Laboratory for Topology and Neuroscience\\
  EPFL, Lausanne, Switzerland\\
  \texttt{gspr@nonempty.org}  
}
\input{common.tex}

\begin{document}

\maketitle

\input{abstract.tex}

\input{introduction.tex}

\input{method.tex}

\input{experiments.tex}

\input{discussion.tex}

\input{acknowledgments.tex}

\printbibliography

\appendix

\input{supplementary.tex}

\end{document}

%% file: common.tex
\theoremstyle{definition}

\theoremstyle{plain}

\definecolor{mplC0}{HTML}{1f77b4}
\definecolor{mplC1}{HTML}{ff7f0e}
\definecolor{mplC2}{HTML}{2ca02c}
\definecolor{mplC3}{HTML}{d62728}
\definecolor{mplC4}{HTML}{9467bd}
\definecolor{mplC5}{HTML}{8c564b}
\definecolor{mplC6}{HTML}{e377c2}
\definecolor{mplC7}{HTML}{7f7f7f}
\definecolor{mplC8}{HTML}{bcbd22}
\definecolor{mplC9}{HTML}{17becf}

\newcommand{\RR}{\ensuremath{\mathbb{R}}}

\newcommand{\lapu}{\ensuremath{\mathcal{L}^{\text{up}}}}
\newcommand{\lapd}{\ensuremath{\mathcal{L}^{\text{down}}}}
\newcommand{\lap}{\ensuremath{\mathcal{L}}}

\newcommand{\inv}{\ensuremath{^{-1}}}
\newcommand{\transpose}{\ensuremath{^{\intercal}}}

\tikzset{twosimp/.style={fill opacity=0.4,fill=blue,draw opacity=1.0}}

\newcommand{\ie}{}
\def\ie/{i.e.}
\newcommand{\eg}{}
\def\eg/{e.g.}
\newcommand{\etc}{}
\def\etc/{etc.}

\newcommand{\ip}[2]{\ensuremath{\left\langle #1 , #2 \right\rangle}}

\DeclareMathOperator{\diag}{diag}

\usepackage{color}
\definecolor{darkorchid}{rgb}{0.6,0.196,0.8}


%% file: abstract.tex
\begin{abstract}
We present simplicial neural networks (SNNs), a generalization of graph neural networks to data that live on a class of topological spaces called simplicial complexes.
These are natural multi-dimensional extensions of graphs that encode not only pairwise relationships but also higher-order interactions between vertices---allowing us to consider richer data, including vector fields and $n$-fold collaboration networks.
We define an appropriate notion of convolution that we leverage to construct the desired convolutional neural networks.
We test the SNNs on the task of imputing missing data on coauthorship complexes.
Code and data are available at \url{https://github.com/stefaniaebli/simplicial_neural_networks}.
\end{abstract}

%% file: introduction.tex
\section{Introduction}

The key to convolutional neural networks (CNNs) lies in the way they employ convolution as a local and shift-invariant operation on Euclidean spaces, e.g.\ $\RR$ for audio or $\RR^2$ for images.
Recently, the concept of CNNs has been extended to more general spaces to exploit different structures that may underlie the data:
This includes spherical convolutions for rotationally invariant data~\cite{cohen2018sphericalcnn,esteves2018sphericalcnn,defferrard2020deepsphere}, more general convolutions on homogeneous spaces~\cite{cohen2016groupnn,kondor2018groupnn,worrall2017harmonicnn}, or convolutions on graphs~\cite{bruna2014graphnn,defferrard2016convolutional}.

Graph neural networks (GNNs) have proven to be an effective tool that can take into account irregular graphs to better learn interactions in the data~\cite{bronstein2017geometric,wu2020survey}.
Although graphs are useful in describing complex systems of irregular relations in a variety of settings, they are intrinsically limited to modeling pairwise relationships. The advance of topological methods in machine learning~\cite{Gabrielsson2020topological, Hofer2019LearningRO, rieck2018neural}, and the earlier establishment of \emph{topological data analysis (TDA)}~\cite{carlsson2008,chazal2017,edelsbrunner2010computational,ghrist2008barcodes} as a field in its own right, have confirmed the usefulness of viewing data as topological spaces in general, or in particular as simplicial complexes. The latter can be thought of as a higher-dimensional analog of graphs~\cite{moore2012,patania2017}. We here take the view that structure is encoded in \emph{simplicial complexes}, and that these represent $n$-fold interactions. In this setting, we present \emph{simplicial neural networks (SNNs)}, a neural network framework that take into account locality of data living over a simplicial complex in the same way a GNN does for graphs or a conventional CNN does for grids.

Higher-order relational learning methods, of which hypergraph neural networks~\cite{feng2018hypergraphs} and motif-based GNNs~\cite{monti2018motif} are examples, have already proven useful in some applications, e.g.\ protein interactions~\cite{ze2020graph}. However the mathematical theory underneath the notion of convolution in these approaches does not have clear connections with the global topological structure of the space in question. This leads us to believe that our method, motivated by Hodge--de Rham theory, is far better suited for situations where topological structure is relevant, such as perhaps in the processing of data that exists naturally as vector fields or data that is sensitive to the space's global structure.

%% file: method.tex
\section{Proposed method}

\input{simpl.tex}

\input{scnn.tex}

%% file: simpl.tex
\paragraph{Simplicial complexes.} 
A simplicial complex is a collection of finite sets closed under taking subsets.
We call a member of a simplicial complex $K$ a \emph{simplex} of \emph{dimension $p$} if it has cardinality $p+1$, and denote the set of all such $p$-simplices $K_p$.
A $p$-simplex has $p+1$ \emph{faces} of dimension $p-1$, namely the subsets omitting one element. We denote these $[v_0,\dotsc,\hat{v}_i,\dotsc, v_p]$ when omitting the $i$'th element.
If a simplex $\sigma$ is a face of $\tau$, we say that $\tau$ is a \emph{coface} of $\sigma$. While this definition is entirely combinatorial, there is a geometric interpretation, and it will make sense to refer to and think of $0$-simplices as \emph{vertices}, $1$-simplices as \emph{edges}, $2$-simplices as \emph{triangles}, $3$-simplices as \emph{tetrahedra}, and so forth (see Figure~\ref{fig:data2complex}, (b)).

Let $C^p(K)$ be the set of functions $K_p\to\RR$, with the obvious vector space structure. These \emph{$p$-cochains} will encode our data. Define the linear \emph{coboundary} maps $\delta^p:C^p(K)\to C^{p+1}(K)$ by
\begin{equation*}
\delta^p(f)([v_0,\dotsc,v_{p+1}]) = \sum_{i=0}^{p+1} (-1)^i f([v_0,\dotsc,\hat{v}_i,\dotsc,v_{p+1}]).
\end{equation*}
Observe that this definition can be thought of in geometric terms: The support of $\delta^p(f)$ is contained in the set of $(p+1)$-simplices that are cofaces of the $p$-simplices that make up the support of $f$.

\begin{figure}[htpb]
\savebox{\tempbox}{
\scriptsize{
\begin{tabular}{llr}
    \toprule
    Papers & Authors & Citations \\
    \midrule
    Paper I & A, B, C  & 100  \\
    Paper II &  A, B & 50\\
    Paper III & A, D & 10\\
    Paper IV & C, D & 4\\
    \bottomrule
  \end{tabular}
}}%
\settowidth{\tempwidth}{\usebox{\tempbox}}%
\hfil\begin{minipage}[b]{\tempwidth}%
\raisebox{-\height}{\usebox{\tempbox}}%
\scriptsize{\caption*{(a)}}%
\label{table:data}%
\end{minipage}%
\savebox{\tempbox}{
\input{figures/cc1.tex}
}%
\settowidth{\tempwidth}{\usebox{\tempbox}}%
\hfil\begin{minipage}[b]{\tempwidth}%
\raisebox{-\height}{\usebox{\tempbox}}%
\vspace{-3pt}
\scriptsize{\captionof*{figure}{(b)}}%
\label{fig:co-authoship-complex}%
\end{minipage}%
\savebox{\tempbox}{\scriptsize{
\begin{blockarray}{cccccc}
\tiny{AB} & \tiny{AC} & \tiny{AD} & \tiny{BC} & \tiny{CD} \\
\begin{block}{(ccccc)c}
  3 & 0 & 1 & 0 & 0 & \tiny{AB} \\
  0 & 3 & 1 & 0 & -1 & \tiny{AC} \\
  1 & 1 & 2 & 0 & 1 & \tiny{AD} \\
  0 & 0 & 0 & 3 & -1 & \tiny{BC}\\
  0 & -1 & 1 & -1 & 2 & \tiny{CD}\\
\end{block}
\end{blockarray}}}%
\settowidth{\tempwidth}{\usebox{\tempbox}}%
\hfil\begin{minipage}[b]{\tempwidth}%
\raisebox{-\height}{\usebox{\tempbox}}%
\vspace{-7pt}
\scriptsize{\captionof*{figure}{(c)}}%
\end{minipage}%

\caption{Constructing a simplicial complex from data. (a)~Coauthorship data. (b)~Coauthorship complex with corresponding cochains from the data. (c)~Degree-$1$ Laplacian $L_1$ of the coauthorship complex.}\label{fig:data2complex}
\end{figure}

\paragraph{Simplicial Laplacians.}
We are in this paper concerned with finite abstract simplicial complexes, although our method is applicable to a much broader setting, e.g.\ CW-complexes. In analogy with Hodge--de Rham theory~\cite{madsen1997calculus}, we define the \emph{degree-$i$ simplicial Laplacian} of a simplicial complex $K$ as the linear map
\begin{align*}
  &\lap_i:C^i(K)\to C^i(K) \\
  &\lap_i = \lapu_i + \lapd_i = \delta^{i\ast}\circ\delta^{i} + \delta^{i-1}\circ\delta^{i-1\ast},
\end{align*}
where $\delta^{i\ast}$ is the adjoint of the coboundary with respect to the inner product (typically the one making the indicator function basis orthonormal). In most practical applications, the coboundary can be represented as a sparse matrix $B_i$ and the Laplacians can be efficiently computed as $L_i=B_i\transpose B_{i}+B_{i-1}B_{i-1}\transpose$. The matrices $L_0$ and $B_0$ are the classic graph Laplacian and incidence matrix. Note that the Laplacians carry valuable topological information about the complex: The kernel of the $k$-Laplacian is isomorphic to the $k$-(co)homology of its associated simplicial complex~\cite{eckmann1944,horak2013spectra}\footnote{In other words, the number of zero-eigenvalues of the $k$-Laplacian corresponds to the number of $k$-dimensional holes in the simplicial complex.}.

%% file: figures/cc1.tex
\begin{tikzpicture}[font=\scriptsize] 
  \coordinate (b) at (0,0);
  \coordinate (c) at (2.7,0); 
  \coordinate (a) at (1.35,1.35); 
  \coordinate (d) at (3.7,0.8); 

  \draw[twosimp] (a) -- (b) -- (c) -- cycle;
  \draw (a) -- (d) -- (c) -- cycle;
  \fill[color=black] (a) circle (1.5pt);
  \fill[color=black] (b) circle (1.5pt);
  \fill[color=black] (c) circle (1.5pt);
  \fill[color=black] (d) circle (1.5pt);

  \draw (a) -- (b) node [midway,sloped,above] {$150$};
  \draw (a) -- (c) node [midway,sloped,above] {$100$};
  \draw (b) -- (c) node [midway,sloped,below] {$100$};
  \draw (c) -- (d) node [midway,sloped,below] {$4$};
  \draw (a) -- (d) node [midway,sloped,above] {$10$};
  \node () at (1.3,0.5) {$100$};
  
  \node[anchor=south west] at (a) {$160$};
  \node[anchor=south east] at (a) {$A$};
  
  \node[anchor=north] at (b) {$150$};
  \node[anchor=south] at (b) {$B$};

  \node[anchor=north] at (c) {$104$};
  \node[anchor=south] at (c) {$C$};

  \node[anchor=north] at (d) {$14$};
  \node[anchor=south] at (d) {$D$};
\end{tikzpicture}

%% file: scnn.tex
\paragraph{Simplicial convolution.}

A convolutional layer is of the form $\psi\circ(f\ast \varphi_W)$, where $\ast$ denotes convolution, $\varphi_W$ is a function
\emph{with small support} parameterized by learnable weights $W$, and $\psi$ is some nonlinearity and bias. This formulation of CNNs lends itself to a spectral interpretation that we exploit to extend CNNs to a much more general setting.

Following~\cite{defferrard2016convolutional} and motivated by the fact that the discrete Fourier transform of a real-valued function on an $n$-dimensional cubical grid coincides with its decomposition into a linear combination of the eigenfunctions of the graph Laplacian for that grid, we define the Fourier transform of real $p$-cochains on a simplicial complex with Laplacians $\mathcal{L}_p$ as
\begin{align*}
  &\mathcal{F}_p: C^p(K) \to \mathbb{R}^{\lvert K_p \rvert} \\
  &\mathcal{F}_p(c) = \left(\ip{c}{e_1}_p, \ip{c}{e_2}_p, \dotsc, \ip{c}{e_{\lvert K_p \rvert}}_p\right),
\end{align*}
where the $e_i$'s are the eigencochains of $\mathcal{L}_p$ ordered by eigenvalues $\lambda_1\leq\dotsm\leq\lambda_{\lvert K_p \rvert}$. The function $\mathcal{F}_p$ is invertible since $\mathcal{L}_p$ is diagonalizable; explicitly, if we write $U\diag(\Lambda)U\transpose$ for a normalized eigendecomposition, the orthonormal matrices $U$ and $U\transpose$ represent $\mathcal{F}\inv_p$ and $\mathcal{F}_p$, respectively. This is the foundation for Barbarossa's development of signal processing on simplicial complexes~\cite{barbarossa2018learning}.

Recall that on the function classes for which it is defined, the classical Fourier transform satisfies $\mathcal{F}(f\ast g)=\mathcal{F}(f)\mathcal{F}(g)$, where the right-hand side denotes pointwise multiplication. This will be our definition of convolution in the simplicial setting. Indeed, for cochains $c,c'\in C^p(K)$ we \emph{define} their convolution as the cochain
\begin{equation*}
  c\ast_p c' = \mathcal{F}_p\inv\left(\mathcal{F}_p(c)\mathcal{F}_p(c')\right).
\end{equation*}

Within this framework, we are led to define a \emph{simplicial convolutional layer} with input $p$-cochain $c$ and weights $W$ as being of the form
\begin{equation*}
  \psi\circ\left(\mathcal{F}\inv_p(\varphi_W)\ast_p c\right)
  \end{equation*}
for some as of yet unspecified $\varphi_W\in\mathbb{R}^{\lvert K_p \rvert}$. To ensure the central property that a convolutional layer be localizing, we demand that $\varphi_W$ be a low-degree polynomial in $\Lambda=(\lambda_1, \dotsc, \lambda_{\lvert K_p \rvert})$, namely
\begin{equation*}
  \varphi_W = \sum_{i=0}^N W_i\Lambda^i = \sum_{i=0}^N W_i(\lambda^i_1, \lambda^i_2, \dotsc, \lambda^i_{\lvert K_p \rvert}),
\end{equation*}
for small $N$. In signal processing parlance, one would say that such a convolutional layer \emph{learns filters that are low-degree polynomials in the frequency domain}.

The reason for restricting the filters to be these low-degree polynomials is best appreciated when writing out the convolutional layer in a basis. Let $L^i_p$ denote the $i$'th power of the matrix for $\mathcal{L}_p$ in, say, the standard basis for $C^p(K)$, and similarly for $c$. Then (ignoring $\psi$),
\begin{equation*}
  \mathcal{F}\inv_p(\varphi_W)\ast_p c = \sum_{i=0}^N W_iU\diag(\Lambda^i)U\transpose c = \sum_{i=0}^N W_i\left(U\diag(\Lambda)U\transpose\right)^i c = \sum_{i=0}^NW_iL^i_pc. 
\end{equation*}
This is important for three reasons, like for traditional CNNs.
First, the convolution can be efficiently implemented by $N$ sparse matrix-vector multiplications: This reduces the computational complexity from $\mathcal{O}(\lvert K_p\rvert^2)$ to $\mathcal{O}(\xi\lvert K_p\rvert)$ where $\xi$ is the density factor.
Second, the number of weights to be learned is reduced from $\mathcal{O}(\lvert K_p\rvert)$ to $\mathcal{O}(1)$.
Third, the operation is $N$-localizing in the sense that if two simplices $\sigma,\tau$ are more than $N$ hops apart, then a degree-$N$ convolutional layer does not cause interaction between $c(\sigma)$ and $c(\tau)$ in its output (see the supplementary material).
Those local interactions (in the spatial domain) can be interpreted as message-passing between simplices.


%% file: experiments.tex
\section{Experimental results}

As many real-world datasets contain missing values, missing data imputation is an important problem in statistics and machine learning~\cite{little1986statistical, nelwamondo2007missing}.
Leveraging the structure underlying the data, GNNs have recently proved to be a powerful tool for this task~\cite{spinelli2020neural}.
Extending this view to higher-dimensional structure, we evaluate the performance of SNNs in imputing missing data over simplicial complexes.

\paragraph{Data.}
A \emph{coauthorship complex (CC)}~\cite{patania2017} is a simplicial complex where a paper with $k$ authors is represented by a $(k-1)$-simplex.
The added subsimplices of the $(k-1)$-simplex are interpreted as collaborations among subsets of authors---a natural hierarchical representation that would be missed by the hypergraph representation of papers as hyperedges between authors.
In general, a simplicial complex representing $n$-fold interactions (e.g.\ between authors) can be constructed as the one-mode projection of a multipartite graph (e.g.\ a paper-author bipartite graph).
The $(k-1)$-cochains are given by the number of citations attributed to the given collaborations of $k$ authors.
See Figure~\ref{fig:data2complex} and~\ref{fig:bipartite}, and the supplementary material for details.
We sampled (see the supplementary material) two coauthorship complexes---CC1 and CC2, see Table~\ref{table:Simplices-coauthor} for statistics---from the Semantic Scholar Open Research Corpus~\cite{ammar18NAACL}, a dataset of over $39$ million research papers with authors and citations.

\paragraph{Method.}
We evaluated the performance of the SNNs on the task of imputing missing citations on the $k$-cochains (for $k=0,1,2$) of the extracted coauthorship complexes.
As in a typical pipeline for this task~\cite{nelwamondo2007missing}, missing values are artificially introduced by replacing a portion of the values with a constant.
Specifically, given a fixed coauthorship complex, missing data is introduced at random on the $k$-cochains at $5$ rates: $10\%, 20\%,  30\% ,40\%$, and $50\%$.
The SNN is given as input the $k$-cochains on which missing citations are substituted by the median of known citations (as a reasonable first guess) and is trained to minimize the $L_1$ norm over known citations.
We trained SNNs made of $3$ layers with $30$ convolutional filters of degree $N=5$ with Leaky ReLu for $1000$ iterations with the Adam optimizer and a learning rate of $10^{-3}$.

\paragraph{Results.}
Figure~\ref{fig:accuracy-error} shows the mean accuracy and absolute error distribution (see the supplementary material for definitions) of the SNN in inputing missing citations on CC1. Observe that the distribution of the prediction error accumulates close to zero.

\begin{figure}[htbp]
  \centering
  \includegraphics[height=2.7cm]{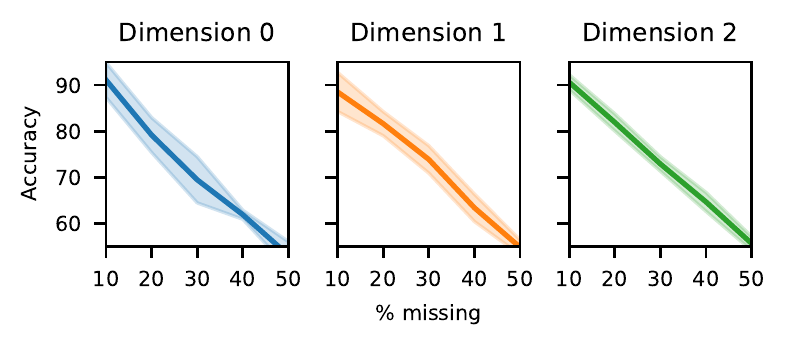}
  \hfill
  \includegraphics[height=2.7cm]{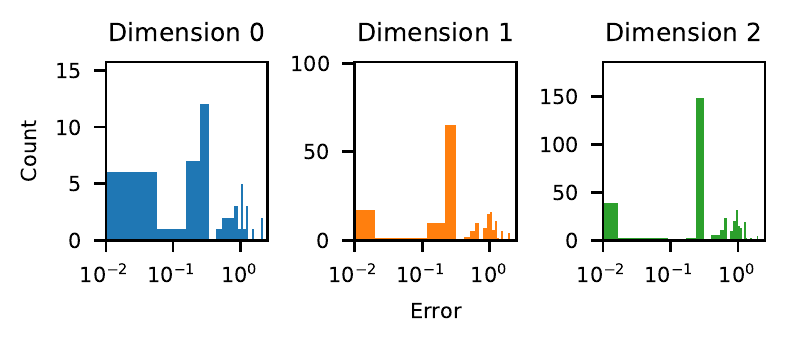}
  \caption{%
    Performance of SNNs.
    Left: Mean accuracy $\pm$ standard deviation over 5 samples in imputing missing citations on CC1.
    Right: Absolute error distribution over 1 sample for $40\%$ missing citations on CC1.
  }\label{fig:accuracy-error}
\end{figure}

Table~\ref{tab:baselines} shows the performance of two baselines: missing values inferred as (i) the mean or median of all known values, and (ii) the mean of the $(k-1)$ and $(k+1)$ neighboring simplices.
SNNs well outperform these baselines.
Comparison with stronger imputation algorithms is left for future work.

To demonstrate that our filters transfer across complexes, we evaluated how accurately an SNN trained on one coauthorship complex can impute missing citations on a different complex.
We found that when imputing citations on CC1, a SNN trained on CC2 is almost as good as one trained on CC1 (compare Figures~\ref{fig:accuracy-error} and~\ref{fig:transfer-learning}).
We expect this result as coauthorship complexes share a similar structure, and the same process underlies the generation of citations across coauthorship complexes.

\begin{table}[htbp]
  \centering
  \begin{minipage}[b]{0.35\linewidth}
    \includegraphics[width=\linewidth]{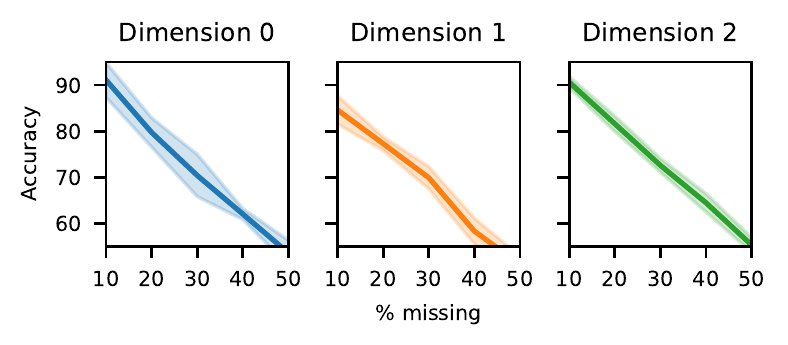}
    \captionof{figure}{Performance on CC1 with an SNN trained on CC2.}\label{fig:transfer-learning}
  \end{minipage}
  \hspace{0.5cm}
  \begin{minipage}[b]{0.56\linewidth}
    \scriptsize{\begin{tabular}{lrrr}
      \toprule
      Method & Dimension 0 & Dimension 1 & Dimension 2 \\
      \midrule
      Global Mean & $3.30\pm0.82$ & $5.75\pm1.28$ & $2.96\pm0.49$ \\
      Global Median & $7.78\pm2.70$ & $10.44\pm1.00$ & $12.50\pm0.63$ \\
      Neighbors Mean & $11.88\pm5.29$ & $24.15\pm1.85$ & $27.38\pm1.18$ \\
      \bottomrule
    \end{tabular}}
    \vspace{9pt}
	\caption{Performance of baselines: mean accuracy $\pm$ standard deviation over 5 samples for $30\%$ missing citations on CC1.}\label{tab:baselines}
  \end{minipage}
\end{table}

%% file: discussion.tex
\section{Conclusion and future work}

We introduced a mathematical framework to design neural networks for data that live on simplicial complexes and provided preliminary results on their ability to impute missing data.
Future work might include:
(i) comparing SNNs with state-of-the-art imputation algorithms,
(ii) using SNNs to solve vector field problems,
(iii) generalizing coarsening and pooling to simplicial complexes,
(iv) using boundaries and coboundaries to mix data structured by relationships of different dimensions,
and (v) studying the expressive power of SNNs.

Unrelated to the simplicial nature of this work, we would like to emphasize how the spectral language was key to developing and even formulating our method.
On homogeneous spaces, convolutions are defined as inner-products with filters shifted by the actions of a symmetry group of the space.
They are the most general shift-invariant linear operators.
On non-homogeneous spaces however, the spectral language yields generalized convolutions which are inner-products with \emph{localized} filters~\cite[Sec.~2.4]{perraudin2019deepsphere}. 
Those too are invariant to any symmetry the space might have.
Convolutions exploit the space's structure to reduce learning complexity by sharing learnable weights through shifts and localizations of filters.

%% file: acknowledgments.tex
\begin{ack}

  S.E.\ and G.S.\ were supported by the Swiss National Science Foundation grant number 200021\_172636, and would like to thank K.\ Hess for valuable discussions.

\end{ack}

%% file: supplementary.tex
\section{Supplementary material}

\paragraph{Simplicial distance and the localizing property of the Laplacian.}
Suppose that $\sigma$ and $\tau$ are $p$-simplices for which $(\nu_0, \nu_1, \dotsc, \nu_d)$ is the shortest sequence of $p$-simplices with the property that $\nu_0=\sigma$, $\nu_d=\tau$, and each $\nu_i$ shares a face or a coface with $\nu_{i-1}$, and a face or a coface with $\nu_{i+1}$. We say that $d$ is the \emph{simplicial distance} between $\sigma$ and $\tau$. Then for all $N<d$, the entry of $L_p^N$ corresponding to $\sigma$ and $\tau$ is $0$, and so the filter does not cause interaction between $c(\sigma)$ and $c(\tau)$. This is analogous to a size-$d$ ordinary CNN layer not distributing information between pixels that are more than $d$ pixels apart. We will refer to $N$ as the \emph{degree} of the convolutional layer, but one may well wish to keep in mind the notion of \emph{size} from traditional CNNs.

\paragraph{Simplicial complexes as the projections of bipartite graphs.}
Given a bipartite graph $X$-$Y$, the simplicial projection on $Y$ is the simplicial complex whose $(k-1)$-simplices are the sets of $k$ vertices in $Y$ that have at least one common neighbor in $X$.
Cochains on the simplicial projection are naturally given by weights on $X$: Given any $(k-1)$-simplex $[y_1,\dots,y_k]$ and its neighboring vertices $\{x_1,\dots,x_j\}\subseteq X$, one can define a $(k-1)$-cochain as $\phi(\{x_1,\dots,x_j\})$, for any function $\phi: \mathcal{P}(X)\to\RR$.
In our coauthorship application, $\phi$ is the sum and the weight of a paper is the number of times it is cited.
See Figure~\ref{fig:bipartite}.

\begin{figure}[htpb]
\savebox{\tempbox}{

  \begin{tikzpicture}[font=\scriptsize]
    \coordinate (I) at (0,0);
    \coordinate (II) at (0,-0.5);
    \coordinate (III) at (0,-1);
    \coordinate (IV) at (0,-1.5);

    \coordinate (A) at ($ (I) + (2,0) $);
    \coordinate (B) at ($ (II) + (2,0) $);
    \coordinate (C) at ($ (III) + (2,0) $);
    \coordinate (D) at ($ (IV) + (2,0) $);

    \fill[color=black] (I) circle (2pt);
    \fill[color=black] (II) circle (2pt);
    \fill[color=black] (III) circle (2pt);
    \fill[color=black] (IV) circle (2pt);
    \fill[color=black] (A) circle (2pt);
    \fill[color=black] (B) circle (2pt);
    \fill[color=black] (C) circle (2pt);
    \fill[color=black] (D) circle (2pt);

    \node[anchor=east] at (I) {Paper \MakeUppercase{\romannumeral 1}, $100$ citations};
    \node[anchor=east] at (II) {Paper \MakeUppercase{\romannumeral 2}, $50$ citations};
    \node[anchor=east] at (III) {Paper \MakeUppercase{\romannumeral 3}, $10$ citations};
    \node[anchor=east] at (IV) {Paper \MakeUppercase{\romannumeral 4}, $4$ citations};

    \node[anchor=west] at (A) {Author $A$};
    \node[anchor=west] at (B) {Author $B$};
    \node[anchor=west] at (C) {Author $C$};
    \node[anchor=west] at (D) {Author $D$};

    \draw (I) -- (A);
    \draw (I) -- (B);
    \draw (I) -- (C);

    \draw (II) -- (A);
    \draw (II) -- (B);

    \draw (III) -- (A);
    \draw (III) -- (D);

    \draw (IV) -- (C);
    \draw (IV) -- (D);
  \end{tikzpicture}
  
}%
\settowidth{\tempwidth}{\usebox{\tempbox}}%
\hfil\begin{minipage}[b]{\tempwidth}%
\raisebox{-\height}{\usebox{\tempbox}}%
\scriptsize{\caption*{(a)}}%
\end{minipage}%
\savebox{\tempbox}{

  \begin{tikzpicture}[font=\scriptsize]
    \coordinate (A) at (0,0);
    \coordinate (B) at (2,0);

    \coordinate (C) at ($ (A) + (300:1) $);
    \node[anchor=north] at (C) {$100$};

    \coordinate (D) at ($ (B) + (240:1) $);
    \node[anchor=north] at (D) {$50$};

    \draw[dashed] (A) -- (C) -- (B);
    \draw[dashed] (A) -- (D) -- (B);

    \coordinate (AA) at ($ (A) + (0, -1.5) $);
    \coordinate (BB) at ($ (B) + (0, -1.5) $);
   
    \fill[color=black] (A) circle (2pt);
    \fill[color=black] (B) circle (2pt);
    \fill[color=black] (AA) circle (2pt);
    \fill[color=black] (BB) circle (2pt);

    \draw (A) -- (B);
    \draw (AA) -- node[below, align=center] {$1$-cochain\\ $150=100+50$} (BB);

    \node[anchor=east] at (A) {$A$};
    \node[anchor=west] at (B) {$B$};
    \node[anchor=east] at (AA) {$A$};
    \node[anchor=west] at (BB) {$B$};
 
  \end{tikzpicture}
}%
\settowidth{\tempwidth}{\usebox{\tempbox}}%
\hfil\begin{minipage}[b]{\tempwidth}%
\raisebox{-\height}{\usebox{\tempbox}}%
\scriptsize{\captionof*{figure}{(b)}}%
\end{minipage}%
\vspace{5pt}
\savebox{\tempbox}{
\input{figures/cc1.tex}
}
\settowidth{\tempwidth}{\usebox{\tempbox}}%
\hfil\begin{minipage}[b]{\tempwidth}%
\raisebox{-\height}{\usebox{\tempbox}}%
\scriptsize{\captionof*{figure}{(c)}}%
\end{minipage}%
\caption{%
    Constructing a simplicial complex and its cochain from a bipartite graph.
    (a)~Paper-author bipartite graph (same data as in Figure~\ref{fig:data2complex}).
    (b)~The $1$-simplex $[A,B]$ is included in the coauthorship complex since authors $A$ and $B$ collaborated on papers I and II.
    The 1-cochain on $[A,B]$ is given by the sum of their common papers' citations.
    (c)~Resulting coauthorship complex with cochains.
}\label{fig:bipartite}
\end{figure}

\begin{table}[htbp]
  \centering
  \scriptsize{
  \begin{tabular}{lrrrrrrrrrrr}
    \toprule
    Dimension   & 0     & 1  & 2     & 3 & 4     & 5 & 6    & 7 & 8   & 9 & 10\\
    \midrule
    CC1 & 352  & 1474  & 3285  & 5019  & 5559  & 4547  & 2732  & 1175  & 343 & 61 & 5\\
    CC2 & 1126 & 5059 & 11840 & 18822 & 21472 & 17896  & 10847 & 4673 & 1357 & 238 & 19\\
    \bottomrule
  \end{tabular}}
  \vspace{2pt}
  \caption{%
  Number of simplices of the two coauthorship complexes sampled from Semantic Scholar.
  } \label{table:Simplices-coauthor}
\end{table}

\paragraph{Sampling papers.}
From the Semantic Scholar Open Research Corpus~\cite{ammar18NAACL}, we excluded papers with fewer than $5$ citations or more than $10$ authors.
To sample a CC, we sampled $80$ papers (corresponding to maximal simplices in the CC) by performing a random walk (of length $80$, from a randomly chosen starting paper) on the graph whose vertices represent papers and edges connect papers sharing at least one author.

\paragraph{Mean accuracy and absolute error.}
A missing value is considered to be correctly imputed if the imputed value differs by at most $10\%$ from the true value.
The \emph{accuracy} is the percentage of missing values that has been correctly imputed and the \emph{absolute error} is the magnitude of the difference between the imputed and true value.
For each rate of missing values, we compute the \emph{mean accuracy} $\pm$ standard deviation over 5 samples with randomly damaged portions.